\documentclass{article}
\usepackage{ijcai17}

\usepackage{times}

\usepackage{latexsym}
\usepackage{comment}
\usepackage{times}
\usepackage[utf8x]{inputenc}
\usepackage[T1]{fontenc}
\usepackage{graphicx}
\usepackage{comment}
\usepackage{textcomp}
\usepackage{float}
\usepackage{subfigure}
\usepackage{setspace}
\usepackage{comment}
\usepackage{amssymb}
\usepackage{setspace}

\title{Effective Spoken Language Labeling with Deep Recurrent Neural Networks}

\author{Marco Dinarelli, Yoann Dupont, Isabelle Tellier\\
LaTTiCe (UMR 8094), CNRS, ENS Paris, Universit\'e Sorbonne Nouvelle - Paris 3\\
PSL Research University, USPC (Universit\'e Sorbonne Paris Cit\'e)\\
1 rue Maurice Arnoux, 92120 Montrouge, France\\
marco.dinarelli@ens.fr, yoa.dupont@gmail.com, isabelle.tellier@univ-paris3.fr}

\begin{document}

\maketitle

\begin{abstract}
Understanding spoken language is a highly complex problem, which can be decomposed into several simpler tasks. In this paper, we focus on Spoken Language Understanding (SLU), the module of spoken dialog systems responsible for extracting a semantic interpretation from the user utterance. The task is treated as a labeling problem.
In the past, SLU has been performed with a wide variety of probabilistic models.
The rise of neural networks, in the last couple of years, has opened new interesting research directions in this domain.
Recurrent Neural Networks (RNNs) in particular are able not only  to represent several pieces of information  as embeddings
but also, thanks to their recurrent architecture, to encode as embeddings relatively long contexts.
Such long contexts are in general out of reach for models previously used for SLU.
In this paper we propose novel RNNs architectures for SLU which outperform previous ones.
Starting from a published idea as base block, we design new deep RNNs achieving state-of-the-art results on two widely used corpora for SLU: ATIS (Air Traveling Information System), in English, and MEDIA (Hotel information and reservation in France), in French.
\end{abstract}

\section{Introduction}

One of the most important step towards building intelligent machines is allowing humans and computers to interact using spoken language.
This task is very hard. As a first approximation thus, spoken dialog system applications have been designed where humans can interact with computers on a specific domain.
In this context, effective human computer interactions depend on the Spoken Language Understanding (SLU) module of a spoken dialog system \cite{demori08-SPM}, 
which is responsible for extracting a semantic interpretation from the user utterance.
A correct interpretation is crucial, as it allows the system to correctly understand the user will, 
to correctly generate the next dialog turn and in turn to achieve a more human-like interaction.
In the past, SLU modules have been designed with a wide variety of probabilistic models \cite{Gupta2006:ATTSLU,raymond07-luna,Hahn.etAL-SLUJournal-2010,Dinarelli.etAl-SLU-RR-2011}.
The rise of neural networks, in the last couple of years, has opened new interesting research directions in this domain \cite{RNNforSLU-Interspeech-2013,Vukotic.etal_2015,Vukotic.etal_2016}.
Recurrent Neural Networks \cite{jordan-serial,werbos-bptt,Cho-2014-GatedRecurrentUnits,LeakyReLU-PReLU-2015} seem particularly adapted to this task.
They allow not only to represent several pieces of information as embeddings but also, thanks to their recurrent architecture, to encode as embeddings relatively long contexts.
This is a very important feature in spoken dialog systems, as the correct interpretation of a dialog turn may depend on the information extracted from previous turns.
Such long contexts are in general out of reach for models previously used for SLU.

\begin{figure}
  \centering
  \subfigure[Jordan]{
    \includegraphics[height=3.25cm]{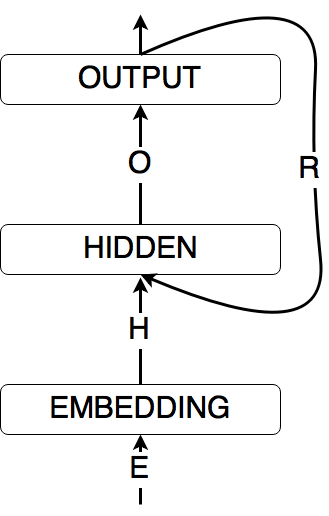}
  }
  \quad
  \subfigure[I-RNN variant]{
    \includegraphics[height=3.25cm]{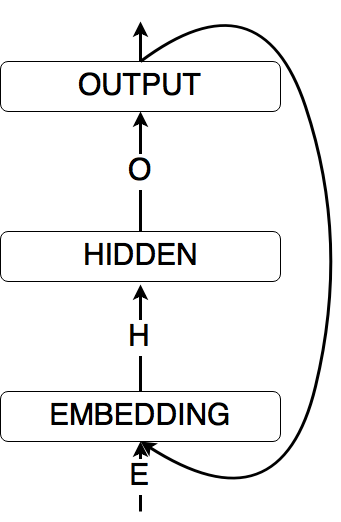}
  }
  \vspace{-0.8em}
  \caption{\scriptsize{Jordan RNN and \emph{I-RNN} variant used in this paper.}}
	\label{fig:3architecgtures}
        \vspace{-1.0em}
\end{figure}

We propose novel deep Recurrent Neural Networks for SLU, treated as a sequence labeling problem.
In this kind of tasks, effective models can be designed by learning \emph{label dependencies}.
For this reason, we start from the idea of \emph{I-RNN} in \cite{DinarelliTellier:RNN:CICling2016}, which uses label embeddings together with word embeddings to learn label dependencies.
Output labels are converted into label indexes and given back as inputs to the network,
they are thus mapped into embeddings the same way as words.
Ideally, this kind of RNN can be seen as an extension of the simple Jordan model \cite{jordan-serial}, 
where the recurrent connection is a loop from the output to the input layer.
A high level schema of these networks is shown in figure~\ref{fig:3architecgtures}.

In this paper we capitalize from previous work described in \cite{DinarelliTellier:RNN:CICling2016,2016:arXiv:DinarelliTellier:NewRNN,Dupont.etAl:LDRNN:CICling2017}.
We use the \emph{I-RNN} of \cite{DinarelliTellier:RNN:CICling2016} as base block to design more effective, 
deep RNNs.
We propose in particular two new architectures.
In the first one, the simple \emph{ReLU} hidden layer is replaced by a \emph{GRU} hidden layer \cite{Cho-2014-GatedRecurrentUnits}, which has proved to be able to learn long contexts.
In the second one, we take advantage of deep networks, by using two different hidden layers: (i)
the first level is split into different hidden layers, one for each type of input information (words, labels and others) in order to learn independent internal representations for each input type; (ii) the second level takes the concatenation of all the previous hidden layers as input, and outputs a new internal representation which is finally used at the output layer to predict the next label.

In particular our \textit{deep} architecture, can be compared to hybrid \textit{LSTM+CRF} architectures proposed in the last years in a couple of papers \cite{huang2015bidirectional,lample2016neural,Ma-Hovy-ACL-2016}.
Such models replace the traditional local decision function of RNNs (the softmax) by a CRF neural layer in order to deal with sequence labeling problems.
Our intuition is that, if RNNs are able to \textit{remember} arbitrary long contexts, by using label information as context they are able to predict correct label sequences without the need of adding the complexity of a neural CRF layer.
In this paper we simply use label embeddings to encode a large label context.
While we don't compare our models on the same tasks as those used in \cite{huang2015bidirectional,lample2016neural,Ma-Hovy-ACL-2016}\footnote{Since we don't have a graphic card, our networks are still relatively expensive to train on corpora like the Penn Treebank.}, 
we compare to LSTM, GRU and traditional CRF models heavily tuned on the same tasks as those we use for evaluation.
Such comparison provides evidence that our solution is a good alternative to complex models like the bidirectional \textit{LSTM+CRF} architecture of \cite{lample2016neural}, as it achieves outstanding performances while being much simpler. Still the two solutions are not mutually exclusive, and their combination could possibly lead to even more sophisticated models.

We evaluate all our models on two SLU tasks: \emph{ATIS} \cite{Dahl-1994-ESA-1075812.1075823}, in English, and \emph{MEDIA}  \cite{Bonneau-Maynard2006-media}, in French.
By combining the use of \emph{label embeddings} for learning label dependencies, and \emph{deep layers for learning internal sophisticated features}, our models achieve state-of-the-art results on both tasks, outperforming strong published models.

In the rest of the paper, we describe in more details our models and we motivate our choices for RNNs (section~\ref{sec:RNNs}). We then describe the tasks used for evaluation, experimental settings and results (section~\ref{sec:Eval}). We end the paper with some conclusions.

  \vspace{-0.8em}
\section{Recurrent Neural Networks}
\label{sec:RNNs}

In this work we use as base block the \emph{I-RNN} proposed in \cite{DinarelliTellier:RNN:CICling2016}.
A similar idea has been proposed in \cite{Bonadiman-etAl-ItaNER-2016}.
In this RNN labels are mapped into embeddings via a look-up table, the same way as words, as described in \cite{Collobert-2008-UAN-1390156.1390177}.
The network uses a matrix $E_w$ for word embeddings, and a matrix $E_l$ for label embeddings, 
of size $N \times D$ and $|O| \times D$, respectively, where $N$ is the size of the word dictionary, 
$D$ is the size chosen for embeddings, while $|O|$ is the number of labels, which corresponds to the size of the output layer.

In order to effectively learn word interactions and label dependencies, 
a wide context is used on both input types, respectively of size $d_w$ for words, and $d_l$ for labels.
We define $E_w(w_i)$ the embedding of any word $w_i$.
The input on the word-side $W_t$ at time step $t$ is then computed as:

{\centering
$W_t = [E_w(w_{t-d_w}) ... E_w(w_{t}) ... E_w(w_{t+d_w})]$

}

\noindent where $[~]$ is the concatenation of vectors (or matrices in the following sections).
Similarly, $E_l(y_i)$ is the embedding of any predicted label $y_i$, and the label-level input at time $t$ is:

{\centering
$L_t = [E_l(y_{t-d_l+1}) E_l(y_{t-d_l+2}) ... E_l(y_{t-1})]$

}

\noindent which is the concatenation of the vectors representing the $d_l$ previous predicted labels.

The hidden layer activities are computed as:

{\centering
$h_t = \Phi( H [W_t L_t] )$

}

\noindent $\Phi$ is an activation function, which is the \textit{Rectified Linear Function} in the basic version of \emph{I-RNN} \cite{DinarelliTellier:RNN:CICling2016} (here and in the following equations we omit biases to keep equations lighter).
The output of the network is computed with a $softmax$ function:

{\centering
$y_t = softmax(O h_t)$

}

\noindent $y_t$ is the predicted label at the processing time step $t$.

\begin{figure}[t]
	\centering
	\footnotesize

	\includegraphics[height=5.7cm]{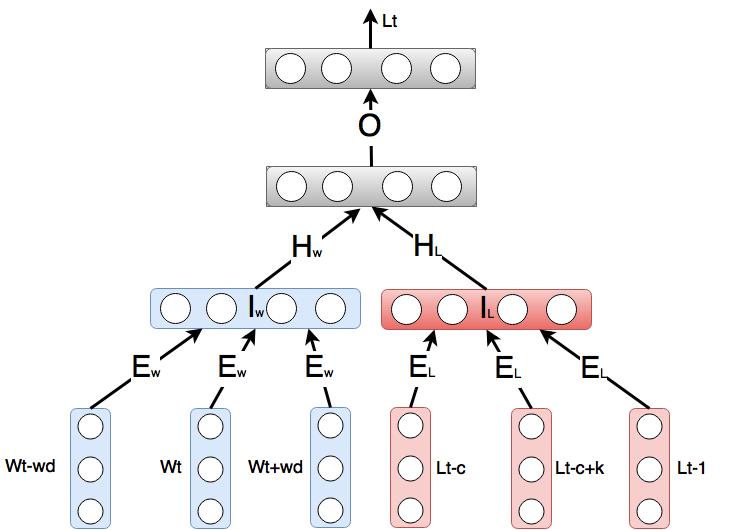}
	  \vspace{-0.8em}
	\caption{\scriptsize{Details of the I-RNN variant used in this paper}}
	\label{fig:LDRNN-details}
	  \vspace{-1.5em}
\end{figure}

A detailed architecture of the \emph{I-RNN} variant used in this work is shown in figure~\ref{fig:LDRNN-details}.
Thanks to the use of label embeddings and to their combination in the hidden layer,
the \emph{I-RNN} variant learns very effectively label dependencies.

\subsection{Deep RNNs}
\label{sec:DeepRNNs}

In this paper we propose two deep RNNs for SLU, which are based on the \emph{I-RNN} variant.

In the first variant, the \emph{ReLU} hidden layer is replaced by a \emph{Gated Recurrent Units} (GRU) hidden layer \cite{Cho-2014-GatedRecurrentUnits},
an improved version of the \emph{LSTM} layer, which proved to be able to learn relatively long contexts.
The architecture of this deep network is the same as the one shown in figure~\ref{fig:LDRNN-details}, 
the only difference is that we use a GRU hidden layer.

\begin{figure}[t]
	\centering
	\footnotesize

	\includegraphics[height=2.5cm]{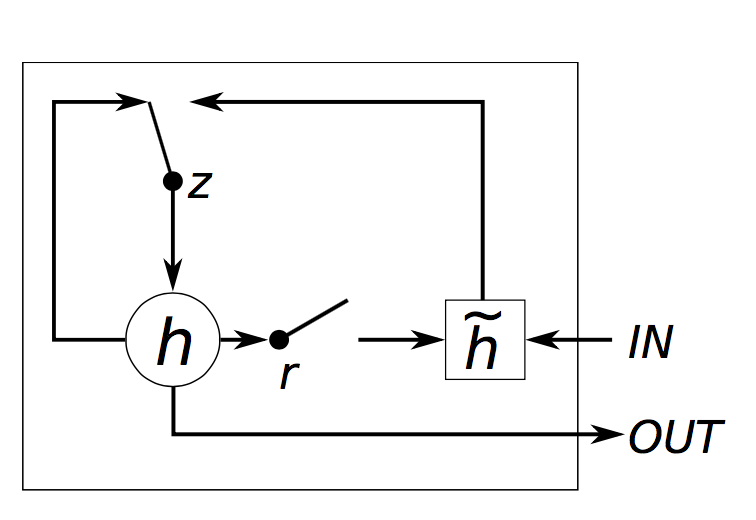}
	  \vspace{-0.8em}
	\caption{\scriptsize{GRU hidden layer, a variant of the LSTM hidden layer.}}
	\label{fig:GRUlayer}
	  \vspace{-0.8em}
\end{figure}

A detailed schema of the GRU hidden layer is shown in figure~\ref{fig:GRUlayer}.
$z$ and $r$ gate units are used to control how past and present information affect the current network prediction.
In particular the $r$ gate learns how to reset past information, making the current decision depends only on current information.
The $z$ gate learns which importance has to be given to current input information.
Combining the two gates and the intermediate value $\hat{h_t}$, the GRU layer can implement the memory cell used in LSTM, which can keep context information for a very long time.
All these steps are computed as follows:

{\centering
$z_t = \Phi( W_z h_{t-1} + U_z W_t )$

$r_t = \Phi( W_r h_{t-1} + U_r W_t )$

$\hat{h_t} = \Gamma( W (r_t \odot h_{t-1}) + U W_t )$

$h_t = (1 - z_t) \odot h_{t-1} + z_t \odot \hat{h_t}$

}

\noindent where $\odot$ is the element-wise multiplication.
In the GRU layer, $\Phi$ is often the \textit{sigmoid} function\footnote{defined as $sigmoid(x) = \frac{1}{1 + e^{-x}}$}, 
while $\Gamma$ is the hyperbolic tangent.\footnote{defined as $tanh(x) = \frac{e^{x} - e^{-x}}{e^{x} + e^{-x}}$}

The second deep RNN proposed in this paper takes advantage of several layers of internal representations.
Deep learning for signal and image processing has shown that several hidden layers allow to learn more and more abstract features \cite{DeepNNForAcousticModeling-Hinton-2012,LeakyReLU-PReLU-2015}.
Such features provide models with a very general representation of information.
While multiple hidden layers have been used also in NLP applications (e.g. \cite{lample2016neural} uses an additional hidden layer on top of a LSTM layer), 
as long as only words are used as inputs, 
it is hard to find an intuitive motivation for using them beyond the empirical evidence that results improve.

\begin{figure}[t]
	\centering
	\footnotesize

	\includegraphics[height=5.7cm]{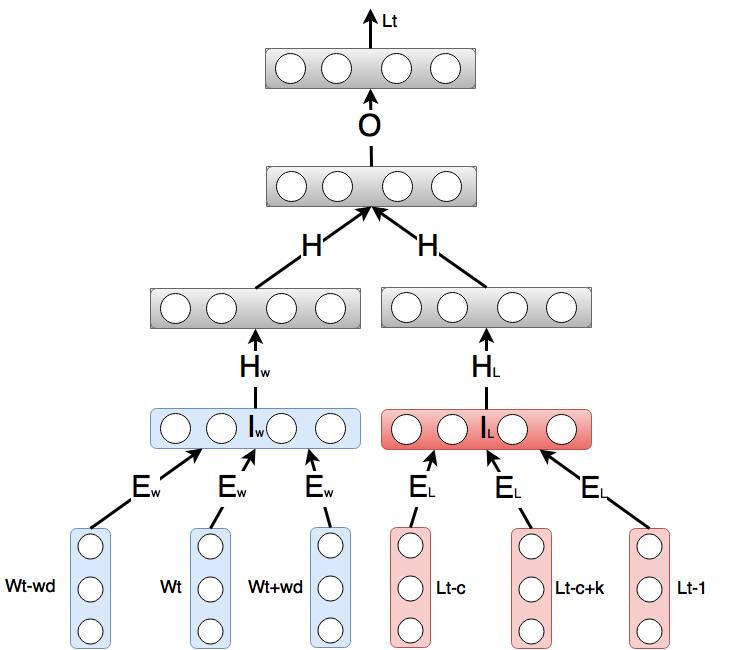}
	  \vspace{-0.8em}
	\caption{\scriptsize{Deep I-RNN proposed in this paper.}}
	\label{fig:Deep-LDRNN-details}
	  \vspace{-0.8em}
\end{figure}

Since the networks described in this paper use in any case at least two different inputs (words and labels), 
the need to learn multiple layers of representations is more clearly justified.
We thus designed a deep RNN architecture where each type of input is connected to its own hidden layer.
In the simplest case, we have one hidden layer for word embeddings and one for label embeddings ($W_t$ and $L_t$ described above). The outputs of both layers are concatenated and given as input to a second global hidden layer.
The output of this second layer is finally processed by the output layer the same way as in the architectures described previously.
A schema of this deep architecture is shown in figure~\ref{fig:Deep-LDRNN-details}.

When other inputs are given to the network (e.g. character-level convolution as described later on), 
in this architecture each of them have its own hidden layer, 
whose outputs are concatenated and given as input to the second hidden layer.

The motivation behind this architecture is that the network learns a different internal representation for each type of input separately in the first hidden layers. Then, in the second hidden layer, the network uses its entire modeling capacity to learn interactions between the different inputs.
With a single hidden layer, the network has to learn both a global internal representation of all inputs and their interactions at the same time, which is much harder.

\subsection{Character-level Convolution Layer}
\label{sec:CharCNN}

Even if word embeddings provide a fine encoding of word features, several works such like \cite{lample2016neural,Ma-Hovy-ACL-2016} have shown that more effective models can be obtained using a convolution layer over the characters of the words.
Character-level information is indeed very useful to allow a model to generalize over rare inflected surface forms and even out-of-vocabulary words in the test phase.
Word embeddings are much less effective in such cases.
Convolution over word characters is even more general, as it can be applied to different languages, allowing to re-use the same system on different languages and tasks.

In this paper we focus on a convolution layer similar to the one used in \cite{Collobert-2011-NLP-1953048.2078186} for words.
For any word $w$ of length $|w|$, we define $E_{ch}(w,i)$ the embedding of the $i$-th character of the word $w$.
We define $W_{ch}$ the matrix of parameters for the linear transformation applied by the convolution (once again we omit the associated vector of biases).
We compute a convolution of window size $2*d_c + 1$ over characters of a word $w$ as follows:

\small
\begin{itemize}
\item $\forall i \in [1,|w|]$
$Conv_i = W_{ch} \cdot [E_{ch}(w,i-d_c); \dots E_{ch}(w,i); \dots E_{ch}(w,i+d_c)]$

\item $Conv_{ch} = [Conv_1 \dots Conv_{|w|}]$

\item $Char_{w} = Max(Conv_{ch})$
\end{itemize}
\normalsize

\noindent the $Max$ function is the so-called max-pooling \cite{Collobert-2011-NLP-1953048.2078186}.
While it is not strictly necessary to map characters into embeddings, it would be probably less interesting applying the convolution on discrete representations.
The matrix $Conv_{ch}$ is made of the concatenation of the vectors returned by the application of the linear transformation. Its size is $|C| \times |w|$, where $|C|$ is the size of the convolution layer.
The max-pooling computes the maxima over the word-length direction, 
thus the final output $Char_{w}$ has size $|C|$, which is independent from the word length.
$Char_{w}$ can be interpreted as a distributional representation of the word $w$ encoding the information at $w$'s character level.
This is a complementary information with respect to word embeddings (which encode inter-word information) and provide the model with an information similar to what is usually brought by discrete lexical features like word prefixes, suffixes, capitalization information etc. and, more in general, with information on the morphology of a language.

\subsection{RNNs Learning}
\label{sec:Learning}

We learn all the networks by minimizing the cross-entropy between the expected label $c_t$ and the predicted label $y_t$ at position $t$ in the sequence, plus a $L2$ regularization term:

{\centering
$C = - c_t \odot log( y_t ) + \frac{\lambda}{2} \left | \Theta \right |^2$

}

\noindent $\lambda$ is a hyper-parameter to be tuned, $\Theta$ stands for all the parameters of the network, 
which depend on the variant used.
$c_t$ is the \textit{one-hot} representation of the expected label.
Since $y_t$ above is the probability distribution over the label set computed by the softmax, 
we can see the output of the network as the probability $P(i | W_t, L_t)$~$\forall i \in [1,m]$, 
where $W_t$ and $L_t$ are the inputs of the network (both words and labels), $i$ is the index of one of the labels defined in the task at hand.

We can thus associate to the \emph{I-RNN} model the following decision function:

{\centering
$ argmax_{i \in [1,m]} P(i | W_t, L_t)$

}

Note that this is a local decision function, as the probability of each label is normalized at each position of a sequence.
Despite this, the use of label-embeddings $L_t$ as context allows the \emph{I-RNN} to effectively model label dependencies.
In contrast, traditional RNNs don't use label embeddings, most of them don't use labels at all, 
their decision function can thus be defined as:

{\centering
$ argmax_{i \in [1,m]} P(i | W_t)$

}

\noindent which can lead to incoherent predicted label sequences.

We use the traditional back-propagation algorithm with momentum to learn our networks \cite{PracticalRecommendations-Bengio-2012}.
Given the recurrent nature of the networks, the Back-Propagation Through Time (BPTT) is often used \cite{werbos-bptt}.
This algorithm consists in unfolding the RNN for $N$ previous steps, $N$ being a parameter to choose, and thus using the $N$ previous inputs and hidden states to update the model's parameters.
The traditional back-propagation algorithm is then applied. This is similar to learning a feed-froward network of depth N.
The BPTT algorithm is supposed to allow the network to learn long contexts.
However \cite{RNNExtensions_Mikolov-ICASSP-2011} has shown that RNNs for language modeling learn best with only $N = 5$ previous steps.
This can be due to the fact that a longer context does not necessarily lead to better performances,
as a longer context is also more noisy.

In this paper we use instead the same strategy as \cite{RNNforSLU-Interspeech-2013}:
we use a wide context of both words and labels, and the traditional back-propagation algorithm.
From the definition of BPTT given above, our solution can be seen as an approximation of the BPTT algorithm.

\vspace{-0.5em}

\subsection{\textit{Forward}, \textit{Backward} and Bidirectional Networks}
\label{subsec:bidir}

The RNNs introduced in this paper are proposed in forward, backward and bidirectional variants \cite{Schuster-1997-BRNN}.
The forward model is what has been described so far.
The architecture of the backward model is exactly the same, 
the only difference being that the backward model processes sequences from the end to the begin.
Labels and hidden layers computed by the backward model can thus be used as future context in a bidirectional model.

Bidirectional models are described in details in \cite{Schuster-1997-BRNN}.
In this paper we use the variant building separate forward and backward models,
and then computing the final output as the geometric mean of the two models:

{\centering
$y_t = \sqrt{y_t^f \odot y_t^b}$

}

\noindent where $y_t^f$ and $y_t^b$ are the output of the forward and backward models, respectively.

\section{Evaluation}
\label{sec:Eval}

\subsection{Tasks for Spoken Language Understanding}
\label{sec:Corpora}

We evaluated our models on two widely used tasks of Spoken Language Understanding (SLU) \cite{demori08-SPM}.

\textbf{The ATIS corpus} (\textit{Air Travel Information System}) \cite{Dahl-1994-ESA-1075812.1075823} was collected for building a spoken dialog system able to provide US flights information.

ATIS is a simple task dating from $1993$.
The training set is made of $4978$ sentences chosen among dependency-free sentences in the \texttt{ATIS-2} and \texttt{ATIS-3} corpora.
The test set is made of $893$ sentences taken from the \texttt{ATIS-3} \texttt{NOV93} and \texttt{DEC94} data.
Since there is no official development set, we took a part of the training set for this purpose.
Word and label dictionaries contain $1117$ and $85$ items, respectively.
We use the version of the corpus published in \cite{raymond07-luna}, 
where some word classes are available as additional model features, such as city names, airport names, time expressions etc.

An example of sentence taken from this corpus is \textit{``I want all the flights from Boston to Philadelphia today''}.
The words \textit{Boston}, \textit{Philadelphia} and \textit{today} are associated to the concepts \emph{DEPARTURE.CITY}, \emph{ARRIVAL.CITY} and \emph{DEPARTURE.DATE}, respectively.
All the other words don't belong to any concept and are associated to the void concept \emph{O} (for Outside).
This example shows the simplicity of this task:
the annotation is sparse, only $3$ words of the sentence are associated to a non-void concept;
there is no segmentation problem, as each concept is associated to exactly one word.

\textbf{The French corpus MEDIA} \cite{Bonneau-Maynard2006-media} was collected to create and evaluate spoken dialog systems providing touristic information about hotels in France.
This corpus is made of $1250$ dialogs which have been manually transcribed and annotated following a rich concept ontology.
Simple semantic components can be combined to create complex semantic structures.
For example the component \emph{localization} can be combined with other components like \texttt{city}, \texttt{relative-distance}, \texttt{generic-relative-location}, \texttt{street} etc.
The MEDIA task is a much more challenging task than ATIS: 
the rich semantic annotation is a source of difficulties, and so is also the annotation of coreference phenomena.
Some words cannot be correctly annotated without taking into account a relatively long context, 
often going beyond a single dialog turn.
For example in the sentence \textit{``Yes, the one which price is less than 50 Euros per night''}, \textit{the one} is a mention of a hotel previously introduced in the dialog.
Moreover labels are segmented over multiple words, creating possibly long label dependencies.

These characteristics, together with the small size of the training data, make MEDIA a much more suitable task for evaluating models for sequence labeling.
Statistics on the corpus MEDIA are shown in table~\ref{tab:MEDIAStats}.

The MEDIA task can be modeled as sequence labeling by chunking the concepts over several words using the traditional \emph{BIO} notation \cite{Ramshaw95-BIO}.
A comparative example of annotation, also showing the word classes available for the two tasks, is shown in the table~\ref{tab:ATIS-MEDIA-exemple}.

The goal of the SLU module is to correctly extract concepts and their normalized values from the surface forms.
The semantic representation used is concise, allowing an automatic spoken dialog system to easily represent the user will. In this paper we focus on concept labeling. The extraction of normalized values  from these concepts can be easily performed with deterministic modules based on rules \cite{Hahn.etAL-SLUJournal-2010}.

\begin{table}[t]
	    \scriptsize
	    \begin{tabular}{|ccc|ccc|}
	    \hline
	    \multicolumn{3}{|c|}{MEDIA} & \multicolumn{3}{|c|}{ATIS} \\
	    \textbf{Words} & \textbf{Classes} & \textbf{Labels} & \textbf{Words} & \textbf{Classes} & \textbf{Labels} \\
	    \hline
        Oui & - & Answer-B & i'd & - & O \\
        l' & - & BDObject-B & like & - & O \\
        hotel & - & BDObject-I & to & - & O \\
        le & - & Object-B & fly & - & O \\
        prix & - & Object-I & Delta & airline & airline-name\\
        à & - & Comp.-payment-B & between & - & O \\
        moins & relative & Comp.-payment-I & Boston & city & fromloc.city\\
        cinquante & tens & Paym.-amount-B & and & - & O \\
        cinq & units & Paym.-amount-I & Chicago & city & toloc.city\\
        euros & currency & Paym.-currency-B & & &\\
        \hline
	    \end{tabular}
	    \caption{\scriptsize{An example of annotated sentence taken from MEDIA (left) and ATIS (right). The translation of the sentence in French is \textit{``Yes, the one which price is less than 50 Euros per night''}}}
	    \label{tab:ATIS-MEDIA-exemple}
    \end{table}

\begin{table}[t]
\begin{minipage}{1.0\linewidth}
    \centering
    \scriptsize
    \begin{tabular}{|l|rr|rr|rr|}
      \hline
      & \multicolumn{2}{|c|}{Training} & \multicolumn{2}{|c|}{Dev.} & \multicolumn{2}{|c|}{Test}\\
      \hline
      \# Sentences     &\multicolumn{2}{|c|}{12,908} &\multicolumn{2}{|c|}{1,259}&\multicolumn{2}{|c|}{3,005} \\
      \hline
      \hline
      & \multicolumn{1}{|c}{words} & \multicolumn{1}{c|}{concepts} &  \multicolumn{1}{|c}{words} & \multicolumn{1}{c|}{concepts} &
      \multicolumn{1}{|c}{words} & \multicolumn{1}{c|}{concepts} \\
      \hline
      \# tokens          & 94,466 & 43,078 & 10,849 & 4,705 & 25,606 & 11,383 \\
      \# vocab.         &  2,210 &     99 &    838 &    66 &  1,276 &     78 \\
      \# OOV\%   & --     & --     &  1.33  & 0.02  &  1.39  &  0.04  \\
      \hline
    \end{tabular}
    \caption{\scriptsize{Statistics of the corpus MEDIA. \# tokens is the number of tokens, \# vocab. is the vocabulary size, \# OOV is the number of Out-of-Vocabulary words.}}
  \label{tab:MEDIAStats}
  \vspace{-2.0em}
  \end{minipage}
\end{table}

\subsection{Settings}
\label{sec:Settings}

All RNNs based on the \emph{I-RNN} are implemented in \textit{Octave}\footnote{https://www.gnu.org/software/octave/; Our code is described at http://www.marcodinarelli.it/software.php and available upon request} using \textit{OpenBLAS} for fast computations..\footnote{http://www.openblas.net; This library allows a speed-up of roughly $330\times$ on a single matrix-matrix multiplication using $16$ cores.}

Our RNN models are trained with the following procedure:

\small
\begin{itemize}
\item Neural Network Language Models (NNLM), like the one described in \cite{Bengio03aneural}, are trained for words and labels to generate the embeddings (separately).
\item Forward and backward models are trained using the word and label embeddings trained at the previous step.
\item The bidirectional model is trained using as starting point the forward and backward models trained at the previous step.
\end{itemize}
\normalsize

The first step is optional, as embeddings can be initialized randomly, or using externally trained embeddings.
Indeed we ran also some experiments using embeddings trained with \textit{word2vec} \cite{Word2Vec_Mikolov-2013}.
The results obtained are not significantly different from those obtained following the procedure described above, 
these results will thus not be given in the following sections

All hyper-parameters and layer sizes of our version of the \emph{I-RNN} variant have been moderately optimized on the development data of the corresponding task.\footnote{Without a graphic card, a full optimization is still relatively expensive.}
The deep RNNs proposed in this paper have been run using the same parameters.
We provide the best values found for the two tasks.

The number of training epochs for both tasks is $30$ for the token-lavel NNLM, $20$ for the label-level NNLM, $30$ for forward and backward taggers, and $8$ for the bidirectional tagger. Since the latter is initialized with the forward and backward models, it is very close to the optimum since the first iteration, it doesn't need thus a lot of learning epochs.
At the end of the training phase, we keep the model giving the best prediction accuracy on the development data.

We initialize all the weights with the \textit{Xavier initialization} \cite{PracticalRecommendations-Bengio-2012}, theoretically motivated in \cite{LeakyReLU-PReLU-2015}.
The initial learning rate is $0.5$, it is linearly decreased during the training phase (\textit{Learing Rate decay}).
We combine \textit{dropout} and $L_2$ regularization \cite{PracticalRecommendations-Bengio-2012}, 
the best value for the dropout probability is $0.5$ at the hidden layer, $0.2$ at the embedding layer on ATIS, $0.15$ on MEDIA. The best coefficient ($\lambda$) for the $L_2$ regularization is $0.01$ for all the models, except for the bidirectional model where the best value is $3e^{-4}$.

The size of the embeddings and of the hidden layer is always $200$, except when all information is used as input (words, labels, classes, character convolution), in which case the hidden layer size is $256$.
The size of character embeddings is always $30$, the size of the convolution layer is $50$ on ATIS, $80$ on MEDIA.
The best size of the convolution window is always $1$, meaning that characters are used individually as input to the convolution.

The best size for word and label contexts are $11$ and $5$ on ATIS, respectively.
$11$ means $5$ words on the left of the current position of the sequence, $5$ on the right, plus the current word, 
while $5$ for the label context means $5$ previous predicted labels.
On MEDIA the best sizes are $7$ and $5$ respectively.

\subsection{Results}
\label{sec:Results}

\begin{table}[!]
    \centering
    \scriptsize
    \begin{tabular}{|l|r|r|r|}
      \hline
      Model & \multicolumn{3}{c|}{F1 measure} \\
        \hline
        \hline
										&	forward	&	backward	&	bidirectional \\
	\hline
	\cite{Vukotic.etal_2016} lstm				&	95.12 		& 	-- 			&	95.23 \\
	\cite{Vukotic.etal_2016} gru					&	95.43 		& 	-- 			&	95.53 \\
	\hline
	\cite{2016:arXiv:DinarelliTellier:NewRNN} E-rnn	&	94.73		&	93.61		&	94.71 \\
	\cite{2016:arXiv:DinarelliTellier:NewRNN} J-rnn	&	94.94		&	94.80		&	94.89 \\
	\cite{2016:arXiv:DinarelliTellier:NewRNN} I-rnn	&	95.21		&	94.64		&	94.75 \\
	\hline
	\hline
	I-$rnn_{GRU}$ Words		&	93.58	&	93.81	&	93.83 \\
	\hline
	\hline
	I-$rnn$ Words			&	94.31	&	94.32	&	94.47 \\
	I-$rnn$ Words+Classes		&	95.37	&	95.44	&	\textbf{95.56} \\	
	I-$rnn$ Words+Classes+CC	&	95.40	&	95.39	&	95.46 \\	
	\hline
	\hline
	I-$rnn_{deep}$ Words				&	94.47	&	94.29	&	94.52 \\	
	I-$rnn_{deep}$ Words+Classes		&	\textbf{95.67}	&	\textbf{95.54}	&	\textbf{95.60} \\	
	I-$rnn_{deep}$ Words+Classes+CC	&	\textbf{95.56}	&	95.39	&	\textbf{95.53} \\	
	\hline
      \hline
    \end{tabular}
    \caption{\scriptsize{Comparison of our results on the ATIS task with the literature, in terms of F1 measure.}}
  \label{tab:SLUATIS}
\vspace{-1.3em}
\end{table}

\begin{table}[!]
    \centering
    \scriptsize
    \begin{tabular}{|l|c|c|c|}
      \hline
      Model & \multicolumn{3}{c|}{F1 measure / Concept Error Rate (CER)} \\
        \hline
        \hline
	&	\multicolumn{1}{c|}{forward}	&	\multicolumn{1}{c|}{backward}	&	\multicolumn{1}{c|}{bidirectional} \\
	\hline
	\cite{Vukotic.etal_2015} CRF				&	\multicolumn{3}{c|}{86.00~/~--} \\
	\cite{Hahn.etAL-SLUJournal-2010} CRF		&	\multicolumn{3}{c|}{--~/~10.6} \\
	\cite{Hahn.etAL-SLUJournal-2010} ROVER$\times6$	&	\multicolumn{3}{c|}{--~/~10.2} \\
	\hline
	\hline
	\cite{Vukotic.etal_2015} E-rnn				&	81.94~/~--		&	--~/~-- 		&	--~/~-- \\
	\cite{Vukotic.etal_2015} J-rnn				&	83.25~/~--		&	--~/~-- 		&	--~/~-- \\
	\hline
	\cite{Vukotic.etal_2016} lstm				&	81.54~/~-- 		& 	--~/~-- 		&	83.07~/~-- \\
	\cite{Vukotic.etal_2016} gru				&	83.18~/~-- 		& 	--~/~-- 		&	83.63~/~-- \\
	\hline
	\cite{2016:arXiv:DinarelliTellier:NewRNN} E-rnn	&	82.64~/~--		&	82.61~/~--		&	83.13~/~-- \\
	\cite{2016:arXiv:DinarelliTellier:NewRNN} J-rnn	&	83.06~/~--		&	83.74~/~--		&	84.29~/~-- \\
	\cite{2016:arXiv:DinarelliTellier:NewRNN} I-rnn	&	84.91~/~--		&	86.28~/~--		&	86.71~/~-- \\
	\hline
	\hline
	I-$rnn_{GRU}$ Words				&	84.50~/~13.56	&	85.87~/~11.92	&	86.33~/~11.38 \\
	\hline
	\hline
	I-$rnn$ Words				&	85.36~/~12.55	&	86.51~/~11.17	&	\textbf{86.98}~/~10.68 \\
	I-$rnn$ Words+Classes		&	85.34~/~12.34	&	86.55~/~10.85	&	\textbf{87.07}~/~10.35 \\
	I-$rnn$ Words+Classes+CC	&	85.47~/~12.23	&	86.70~/~10.69	&	\textbf{87.17}~/~\textbf{10.19} \\
	\hline
	\hline
	I-$rnn_{deep}$ Words			&	85.93~/~11.84	&	\textbf{86.80}~/~10.50	&	\textbf{87.24}~/~\textbf{10.03} \\
	I-$rnn_{deep}$ Words+Classes	&	85.63~/~11.87	&	86.64~/~10.41	&	\textbf{87.30}~/~\textbf{9.83} \\
	I-$rnn_{deep}$ Words+Classes+CC	&	85.73~/~11.81	&	\textbf{86.83}~/~10.33	&	\textbf{87.43}~/~\textbf{9.80} \\
	\hline
    \end{tabular}
    \caption{\scriptsize{Comparison of our results on the MEDIA task with the literature, in terms of F1 measure and Concept Error Rate.}}
  \label{tab:SLUMEDIA}
\vspace{-1.3em}
\end{table}

All results shown in this section are averages over $10$ runs.
Word and label embeddings were learned once for all experiments, for each task.

We provide results obtained with incremental information given as input to the models and made of:
i) Only words (previous labels are always given as input), indicated with \emph{Words} in the tables; ii) words and classes \emph{Words+Classes}; iii) words, classes and character convolution \emph{Words+Classes+CC}.
Our implementation of the \emph{I-RNN} variant is indicated in the tables with I-$rnn$. The version using a GRU hidden layer is indicated with I-$rnn_{GRU}$, while I-$rnn_{deep}$ is the version using two hidden layers, as shown in figure~\ref{fig:Deep-LDRNN-details}. E-rnn and J-rnn are the Elman and Jordan RNNs, respectively, while CRF is the Conditional Random Field model \cite{lafferty01:crf}, which is the best individual model for sequence labeling.

Results obtained on the ATIS task are shown in table~\ref{tab:SLUATIS}.
On this task we compare with \emph{lstm} and \emph{gru} models of \cite{Vukotic.etal_2016}, and with RNNs of \cite{2016:arXiv:DinarelliTellier:NewRNN}. Results in bold are those equal or better than the state-of-the-art, which is the F1 $95.53$ of \cite{Vukotic.etal_2016}.
Note that some works report F1 results over $96$ on the ATIS task, e.g. \cite{Mesnil-RNN-2015}.
However they are obtained on a modified version of the ATIS corpus which makes the task easier.\footnote{This version of the data is associated to the tutorial available at http://deeplearning.net/tutorial/rnnslu.html}.
Since all published works on this task report either F1 measure, or both F1 measure and Concept Error Rate (CER), 
in order to save space we only show results in terms of F1.
We report that the best CER reached with our models is $5.02$, obtained with the forward model \emph{I-$rnn_{deep}$ Words}. To the best of our knowledge this is the best result in terms of CER on this task.

As can be seen in table~\ref{tab:SLUATIS}, all models obtain good results on this task.
As a matter of fact, as mentioned above, this task is relatively simple.
Beyond this, our I-$rnn_{deep}$ network systematically outperforms the other networks, achieving state-of-the-art performances.
Note that, on this task, adding the character-level convolution doesn't improve the results.
We explain this with the fact that word classes available for this task already provide the model with most of the information needed to predict the label. Indeed, results improve by more than one F1 point when using classes compared to those obtained using only words, which are already over $94$. Adding more information as input forces the model to use part of its modeling capacity for associations between character convolution and labels, which may replace correct with wrong associations.

Results obtained on the MEDIA task are shown in table~\ref{tab:SLUMEDIA}.
For this task we compare our results with those of \cite{Vukotic.etal_2015,Vukotic.etal_2016,2016:arXiv:DinarelliTellier:NewRNN,Hahn.etAL-SLUJournal-2010}.
The former obtains the best results in terms of F1, while the latter has, since $2010$, the best results in terms of CER.
Those results are obtained with a combination of $6$ individual models by \emph{ROVER} \cite{fiscus97-rover}, which is indicated in the table with \emph{ROVER$\times6$}.

As mentioned above, this task is much more difficult than ATIS, results in terms of F1 measure are indeed $8$-$12$ points lower. This difficulty is introduced not only by the much richer semantic annotation, but also by the relatively long label dependencies introduced by the segmentation of labels over multiple words.
Not surprisingly thus, the CRF model of \cite{Vukotic.etal_2015} achieves much better performances than traditional RNNs (E-rnn, J-rnn, lstm and gru).
The only model able to outperform CRF is the I-RNN of \cite{2016:arXiv:DinarelliTellier:NewRNN}.
All our RNNs are based on this model, which uses label embeddings the same way as word embeddings.
Label embeddings are pre-trained on reference sequences of labels taken from the training data, and than refined during the training phase of the task at hand.
This allows, in general, to learn first general label dependencies and interactions, based only on their co-occurrences.
In the learning phase then, label embeddings are refined integrating information about their interactions with words.
We observed however, that on small tasks like ATIS and MEDIA, pre-training embeddings doesn't really provide significant improvements. On larger tasks however, learning embeddings increase the performances. We thus keep the pre-training phase as a step of our general learning procedure.
The ability of our variant to learn label-word interactions, together with the ability of RNNs to encode large contexts as embeddings, makes I-RNN a very effective model for sequence labeling and thus for SLU.
Our basic version of I-RNN uses a ReLU hidden layer and the dropout regularization, in contrast with the I-RNN of \cite{2016:arXiv:DinarelliTellier:NewRNN} which uses a sigmoid and only $L_2$ regularization.
This makes our implementation much more effective, as shown in table~\ref{tab:SLUMEDIA}.

As can be seen in table~\ref{tab:SLUMEDIA}, most of our results obtained with the bidirectional models are state-of-the-art (highlighted in bold) in terms of both F1 measure and CER. This is even more impressive as the best CER result in the literature is \emph{ROVER$\times6$} which is a combination of $6$ individual models.

Some of our results on the test set may seem not significantly better than others, 
e.g. \emph{I-$rnn_{deep}$ Words+Classes} compared to \emph{I-$rnn_{deep}$ Words+Classes+CC} in terms of CER.
However, we optimize our models on development data, where the \emph{I-$rnn_{deep}$ Words+Classes+CC} model obtains a significantly better result ($10.33$ vs. $10.20$).
This slight lack of generalization on the test set may suggest that more fine parameter optimizations may lead to even better results.

Results in the tables show that the  \emph{I-$rnn_{GRU}$} model is less effective than the other variants proposed in the paper.
This outcome is similar to the one of \cite{Vukotic.etal_2016}, which obtains worse results than the other RNNs on MEDIA.
Compared to that work, adding label embeddings in our variant allows to reach higher performances.
In contrast to \cite{Vukotic.etal_2016}, our results on ATIS are particularly low even considering that we don't use classes.
An analyses on the training phase revealed that the GRU hidden layer is a very strong learner: this network's best learning rate is lower than the one of other RNNs ($0.1$ vs. $0.25$), but the final cost function on the training set is much lower than the one reached by the other variants.
Since we could not solve this overfitting problem even changing activation function and regularization parameters, 
we conclude that this hidden layer is less effective on these particular tasks.
In future work we will further investigate this direction on different tasks.

Beyond quantitative results, a shallow analysis of the model's output shows that I-$rnn$ networks are really able to learn label dependencies. The superiority of this model on the MEDIA task in particular, is due to the fact that this model never makes segmentation mistakes, that is \emph{BIO} errors.
Since I-$rnn$ still makes mistakes, this means that once a label annotation starts at a given position in a sequence, even if the label is not the correct one, the same label is kept at following positions. \emph{I-rnn} tends to be coherent with previous labeling decisions. This behavior is due to the use of a local decision function which definitely relies on the label embedding context. This doesn't prevent the model from being very effective.
Interestingly, this behavior also suggests that I-$rnn$ could still benefit from a CRF neural layer like those used in \cite{lample2016neural,Ma-Hovy-ACL-2016}. We leave this as future work.

\section{Conclusions}
\label{sec:Conclusions}

In this paper we tackle the Spoken Language Understanding problem with recurrent neural networks.
We use as basic block for our networks a variant of RNN taking advantage of several label embeddings as output-side context.
The decision functions in our models are still local, but this limitation is overcome by the use of label embeddings, 
which proves very effective at learning label dependencies.
We introduced two new task-oriented architectures of deep RNN for SLU: one using a GRU hidden layer in place of the simple ReLU. The other, \emph{Deep}, using two hidden layers: the first learns separate internal representations of different input information; the second learns interactions between different pieces of such information.
The evaluation on two widely used tasks of SLU proves the effectiveness of our idea.
In particular the \emph{Deep} network achieves state-of-the-art results on both tasks.



\bibliographystyle{named}
\footnotesize
\bibliography{Biblio_MD}

\end{document}